\RequirePackage[dvipdfmx]{graphicx}
\RequirePackage[dvipdfmx]{color}

\documentclass[11pt,letterpaper]{article}
\usepackage{ijcnlp2017}
\usepackage{times}
\usepackage{latexsym}
\usepackage{multirow}
\ijcnlpfinalcopy

\title{Deep Learning Paradigm with Transformed Monolingual Word Embeddings for Multilingual Sentiment Analysis}

\author{
  Yujie Lu \and Tatsunori Mori\\
  Graduate School of Environment and Information Sciences\\
   Yokohama National University, Yokohama, 240851, Japan\\
  {\tt \{luyujie, mori\}@forest.eis.ynu.ac.jp}
}

\date{}

\begin{document}

\maketitle

\begin{abstract}
The surge of social media use brings huge demand of multilingual sentiment analysis (MSA) for unveiling cultural difference. So far, traditional methods resorted to machine translation---translating texts in other languages to English, and then adopt the methods once worked in English. However, this paradigm is conditioned by the quality of machine translation. In this paper, we propose a new deep learning paradigm to assimilate the differences between languages for MSA. We first pre-train monolingual word embeddings separately, then map word embeddings in different spaces into a shared embedding space, and then finally train a parameter-sharing deep neural network for MSA. The experimental results show that our paradigm is effective. Especially, our CNN model outperforms a state-of-the-art baseline by around 2.1\% in terms of classification accuracy.
\end{abstract}

\section{Introduction} \label{Introduction}

The prevalence of social media has allowed for the collection of abundant subjective multilingual texts. Twitter is a particularly significant multilingual data source that provides researchers with sufficient opinion pieces on various topics from all over the world. An analysis of these multilingual opinion texts can reveal the cultural variations in public opinions from different areas. Therefore, an efficient multilingual sentiment analysis (MSA) that can process all multilingual texts (mixed monolingual texts) simultaneously is necessary.

There has been substantial research on monolingual sentiment analysis, including sentiment analysis of traditional reviews (product/movie, etc.; \cite{Pang02, Turney02, Pang08}) and tweets (\cite{Agarwal11, Go09, Xiang14, Mukherjee12}). Instead of creating separate models for each language, an MSA should use a single model (with the same parameters for all languages) to process different texts in different languages.

However, compared with monolingual sentiment analysis, the research on MSA has progressed slowly. One of the reasons for this is that there is no benchmark dataset that supports the evaluation of MSA methods (particularly, its cross-language adaptability). As many previous studies have highlighted, open-source sentiment datasets are imbalanced \cite{Mihalcea07, Denecke08, Wan09, Steinberger11}: there are many freely available annotated sentiment corpora for English; however, such corpora for other languages are scarce or even nonexistent. As a compromise, many of the previous multilingual corpora have been built using human/machine translations, which are unrealistic. 

In this study, we used the MDSU corpus as our training/test dataset \cite{Lu17C}. The MDSU corpus contains three distinct languages (i.e., English, Japanese, and Chinese) and four identical international topics (i.e., iPhone 6, Windows 8, Vladimir Putin, and Scottish Independence), with 5,422 tweets in total. The multilinguality of the corpus makes it the most suitable training/test dataset for MSA. 

Moreover, traditional machine learning methods that are effective in monolingual settings are not necessarily effective in multilingual settings, because they usually require heavy language-specific feature engineering that further needs language-specific resources (e.g., polarity lexicons)/tools (e.g., POS taggers and parsers). This prevents the application of many sophisticated monolingual methods to other languages, particularly the minor languages that lack basic NLP tools. Until now, the most typically used methods of MSA have been based on machine translation (MT): first, texts in other languages are translated into English, and then machine-learning methods are developed based on the expanded English texts. 
% \footnote{Even for languages holding these tools, applying monolingual methods in MSA can possibly cause a computational burden \cite{Steinberger11}.}

\begin{figure}[htbp]
 \center
 \includegraphics[bb=0 0 448 409, width=0.8\linewidth]{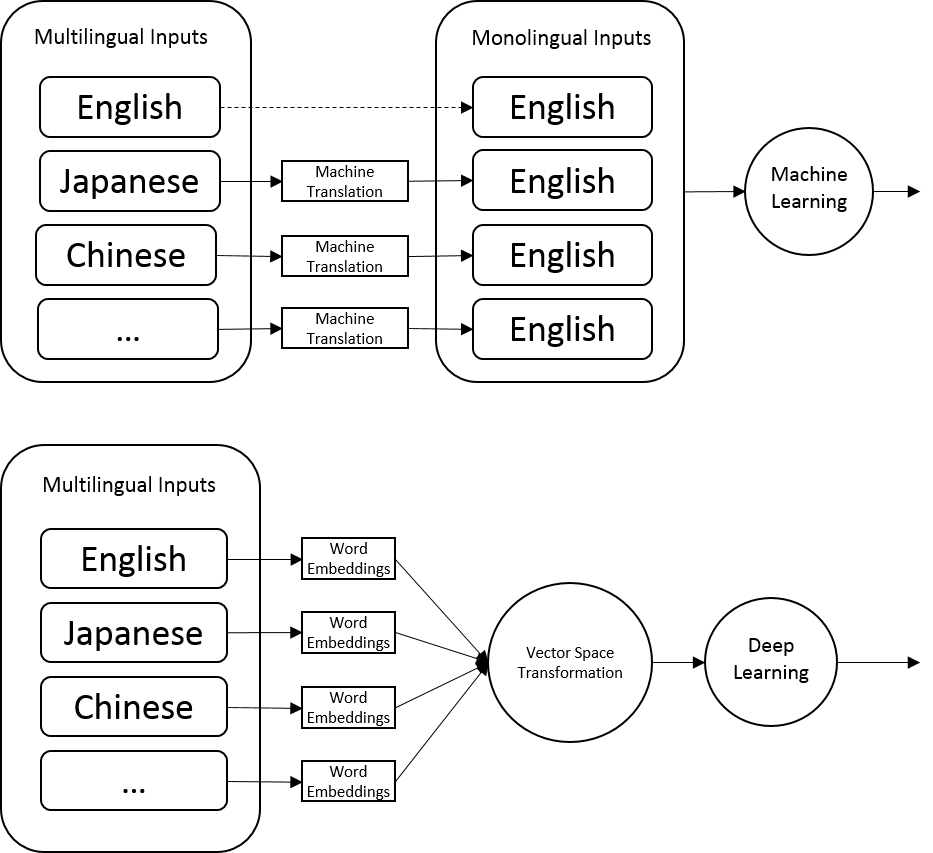}
 \caption{MT-Based Paradigm and Deep Learning Paradigm}
 \label{fig: Paradigms}
\end{figure}

However, this paradigm is conditioned strongly by the quality of the MT. Considering that our processing objects---tweets---contain many informal expressions, it is even more difficult to guarantee an accurate MT. Therefore, we proposed a new deep learning paradigm to integrate the processing of different languages into a unified computation model. First, we pre-trained monolingual word embeddings separately; second, we mapped them in different spaces within a shared embedding space; and finally, we trained a parameter-sharing\footnote{In this thesis, ``parameter-sharing'' specifically means that the same model parameters are shared between different languages.} deep neural network for MSA. Our model is presented in Figure \ref{fig: Paradigms}.

%Figure

Although the study by \cite{Ruder16} is most similar to ours in the use of deep learning methods, there are two fundamental differences. First, they only input the raw monolingual word embeddings (an open-source, pre-trained word embedding for English and random word embeddings for other languages) in their deep learning methods; however, we used customized pre-trained word embeddings and further transferred them into a shared space. Second, they created separate models for each language, whereas we developed a single parameter-sharing model for all languages. 

To the best of our knowledge, this study is the first to use a deep learning paradigm for MSA. Moreover, because of the use of such a paradigm, the only resources we required were word embeddings for each language and tokenizers for non-spaced languages (e.g., Chinese). We expected this paradigm to assimilate language differences to take full advantage of the size of multilingual datasets (compared with its smaller monolingual parts). In this study, we employed the LSTM and CNN models. Our parameter-sharing CNN model with adjusted word embeddings outperformed the machine-translation-based baseline by nearly 5.3\% and the state-of-the-art baseline by 2.1\%, thereby proving its effectiveness.

This paper is organized as follows: in Section \ref{Related Work}, we discuss the related studies; in Section \ref{Methods}, we describe the study methods;  in Section \ref{Experiments}, we presented and discussed  the results of the experiments; and finally, in Section \ref{Conclusion}, we draw conclusions.

\section{Related Work} \label{Related Work}
In this section, we introduce MSA-related studies, including those on multilingual subjectivity analysis as well as the MSA of traditional text and social media.

\subsection{Multilingual Subjectivity Analysis}
Sentiment analysis in a multilingual framework was first conducted for subjectivity analysis. Mihalcea et al. \cite{Mihalcea07} explored the automatic generation of resources (i.e., lexicon translation and corpus projection) for the subjectivity analysis of a new language (i.e., Romanian). They translated the English polarity lexicon into the target language, assessed the quality of the generated lexicon through an annotation study, and proposed a rule-based target-language classifier using the generated lexicon. The results revealed that the translated lexicon was less reliable compared with the English one, and the performance of the rule-based subjectivity classifier was worse in Romanian than in English. They also conducted a subjectivity annotation on a parallel corpus (English sentences were manually translated to Romanian); the results indicated that in most cases, the subjectivity was preserved during the translation. They projected the subjectivity onto the Romanian part to automatically obtain a Romanian subjectivity corpus and trained Naive Bayes (NB) classifiers. The results revealed that the performance of the NB classifiers in Romanian was worse than in English. 

Banea et al. \cite{Banea10} translated the English corpus into other languages (i.e., Romanian, French, English, German, and Spanish) and explored the integration of unigram features from multiple languages into a machine learning approach for subjectivity analysis. They demonstrated that both English and the other languages could benefit from using features from multiple languages. They believed that this was probably because, when one language does not provide sufficient information, another one can serve as a supplement.

\subsection{MSA of Traditional Text}
Although there is extensive scope for improvement, translation-based methods have inspired many other studies. The research on MSA began relatively late. Denecke \cite{Denecke08} translated German movie reviews into English, developed SentiWordNet-based methods for English movie reviews, and tested the proposed methods on the German corpus. The results revealed that the performance of the proposed methods in MSA was similar to that in monolingual settings. Wan \cite{Wan09} leveraged a labeled English corpus for Chinese sentiment classification. He first machine translated the labeled English corpora and an unlabeled Chinese corpus to the target language, and then proposed a co-training approach to use the unlabeled corpora. His experimental results suggested that the co-training approach outperformed the standard inductive and transductive classifiers. Steinberger et al. \cite{Steinberger11} annotated entity-opinion pairs in a parallel news article corpus in seven European languages---English, Spanish, French, German, Czech, Italian, and Hungarian (they first did the annotation work for English and then projected those annotations onto other languages). Their simple method to determine the word polarity aggregation for entity-level sentiment analysis was tested on the entity-opinion pairs in the parallel corpus. They created a valuable resource for entity-level sentiment analysis in a multilingual setting; however, their method, as they observed, is preliminary and depends substantially on language-specific polarity lexicons. 

\subsection{MSA of Social Media}
Recently, the MSA of social media content has been increasing. Balahur and Turchi \cite{Balahur13} conducted an MSA of tweets. They first translated English tweets into four languages---Italian, Spanish, French, and German (the texts in the test set were further corrected manually) to create an artificial multilingual corpus. They then tested support vector machine (SVM) classifiers using polarity lexicon-based features on various combinations of the dataset in different languages. The results suggested that the combined use of training data from multiple languages improves the performance of sentiment classification. Volkova et al. \cite{Volkova13} constructed a multilingual tweet dataset in English, Spanish, and Russian using Amazon Mechanical Turk. They explored the lexical variations in subjective expression and the differences in emoticon and hashtag usage by gender information in the three different languages; their results demonstrated that gender information can be used to improve the performance sentiment analysis of all the three languages.

\subsection{Comparison with Previous Work}
Our study is different from the previous studies in the following ways. First, in multilingual datasets from previous studies, datasets of languages other than English have been projected from the English dataset. Banea et al. \cite{Banea10} and Balahur and Turchi \cite{Balahur13} have used MT to obtain texts in target languages, which are considerably noisy. Mihalcea et al. \cite{Mihalcea07} and Dwnwcke \cite{Denecke08} have directly used parallel corpora to eliminate this noise. However, real multilingual opinion texts would not be in the form of parallel corpora because users usually give their opinions in one language. Therefore, the MDSU corpus in this study includes three distant languages and covers common international topics, which is useful to test the multilingual adaptability of a method.

As for methods, Denecke \cite{Denecke08} and Wan \cite{Wan09} have adopted the ``MT + machine learning'' approach, which unavoidably imports bias during the MT. The abstraction of the word feature in Balahur and Turchi \cite{Balahur13} can be applied to other languages, but it requires language-specific polarity lexicons. Banea et al. \cite{Banea10} used unigrams in multiple languages as features, but they might be restricted due to data sparseness issues. Volkova et al. \cite{Volkova13} proved the effectiveness of employing gender information, but their classifiers are not designed for multilingual settings. By contrast, our deep learning methods require no polarity lexicons and can unify different languages through a neural text model that uses word embeddings. 

%In addition, the novelty of our study is not in the complexity of the network itself\footnote{More sophisticated networks than those used in this study have been proposed for monolingual sentiment analysis.}  but more in the unification of heterogeneous monolingual word embeddings and the parameter-sharing model for multilingual datasets.

\section{Methods} \label{Methods}
In this section, we introduce our baseline methods and the proposed deep learning methods (i.e., transformed word embedding + deep learning). The global polarity of the MDSU corpus has three types: positive, negative, and neutral; therefore, our study is technically a three-way classification task. 
%\footnote{For brevity, positive/negative/neutral are denoted as +/-/= respectively, in Figures 2 and 3.}

\subsection{Baselines}
Our first baseline was MT-based. We used Google Translate\footnote{https://cloud.google.com/translate/} to translate Japanese/Chinese tweets into English. Google Translate is a paid service that supports more than 100 languages at various levels. For Japanese and Chinese, neural MT technology was enabled, providing more reliable translation results for the baselines.

%The SVM is an efficient model for document classification. The SVM basically detects a hyperplane represented by its normal vector {\bf w}, which maximizes the margin between two classes. This search then becomes a constrained optimization problem (i.e., a solved convex quadratic programming problem), and the solution can be written as follows:
%\begin{equation}
%{\bf w} = \sum_{i=1}^{n}\alpha_iy_i{\bf x}_i, \quad \alpha_i\ge 0
%\end{equation}
%where ${\bf x}_i$ are support vectors lying on the class boundaries, $\alpha_i$ are coefficients of the support vectors and $y_i$ are true values, each of which is $\in\{1,-1\}$. SVM can solve non-linear tasks using kernel trick as well.

The SVM-based learning methods with n-gram features, proposed by Pang et al. \cite{Pang02} and Go et al. \cite{Pang02}, have been frequently used as baselines in many monolingual (English) studies. Similar to their settings, we used the default SVM model with a linear kernel and $C = 1$ and fed the binarized unigram/bigram term frequencies as features. The one-vs-one strategy was adopted for multiclass classification. Following the traditional paradigm, the SVM model trained on all translated tweets in the MDSU corpus is our first baseline, denoted as MT-SVM. 
%The models were trained with LibSVM \cite{Chang11} via Python scikit-learn library. 

In addition, we re-implemented Banea et al.'s \cite{Banea10} NB model that uses the cumulation of monolingual unigram features. We fine tuned Banea et al.'s method in two ways: first, we used both unigram and bigram as our features; and second, we used all the features instead of parts of them. We denoted this state-of-the-art baseline that does not use language-specific polarity lexicons as Banea (2010)*.
%\footnote{To our best knowledge, Banea et al. \cite{Banea10} adopted a state-of-the-art method that does not use language-specific polarity lexicons.}

\subsection{Deep Learning Paradigm}
\subsubsection{Space Transformation by Translation Matrix}
Since there is no comparable open source word embeddings learnt from Twitter data for multiple languages, we independently obtained word embeddings using numerous monolingual texts for each language. However, these monolingual word embeddings were heterogeneous in terms of vector space (the meaning of each dimension was different between languages.). Hence, we attempted to reduce the discrepancy between monolingual word embeddings.

This notion was adopted from Mikolov et al. \cite{Mikolov13X}. In their study, they highlighted that the same concepts have similar geometric arrangements in their respective vector spaces. This implies that if the matrix transformation is adequately performed, monolingual word embeddings in heterogeneous spaces can be adjusted to a shared vector space. Thereafter, many other ways to conduct this transformation have been proposed \cite{Ruder17}. Following Mikolov et al. \cite{Mikolov13X}, we used the \textit{Translation Matrix} method---to obtain a linear projection between the languages using a set of pivot word pairs.

Assume a set of word pairs $\{{\bf x}_i, {\bf z}_i\}_{i=1}^n$, where ${\bf x}_i$ and ${\bf z}_i$ are the vector representations of word $i$ in the source and target languages, respectively. We aimed to identify a translation matrix ${\bf W}_{S\rightarrow T}$ that minimized the following object function:

\newcommand{\mymin}{\mathop{\rm minimize}\limits}
\begin{equation}
\mymin_{{\bf W}_{S\rightarrow T}} \ \sum_{i=1}^n || {\bf W}_{S\rightarrow T} {\bf x}_i-{\bf z}_i ||^2
\end{equation}

After ${\bf W}_{S\rightarrow T}$ was identified, we mapped the vocabulary matrix ${\bf Z}$\footnote{A matrix that consists of word embeddings of all words in the training corpus.} of one language space to another by computing ${\bf \hat{Z}} = {\bf Z{\bf W}_{S\rightarrow T}}$. For example, we transferred the Japanese vocabulary matrix to the English vector space using  $\hat{{\bf Z}}_J = {\bf Z}_J {\bf W}_{J\rightarrow E}$.

In this paper, we developed two types of translation matrix: ${\bf W}_{J\rightarrow E}$ and ${\bf W}_{C\rightarrow E}$, to unify our separately pre-trained monolingual word embeddings into a shared one. We selected top $K$ high-frequent word in the English training corpus as our pivot words, translated them into Japanese and Chinese (using Google Translate), and finally obtained the translation matrices using a linear regression algorithm. 

Although the linear projection by the \textit{Translation Matrix} method can be considered as a word-level MT, the space transformation is considerably less expensive than building a full-fledged MT system.

\subsubsection{LSTM}
RNNs have received tremendous attention in the NLP field and been employed to complete many tasks, including predicting words/phrases, speech recognition, image caption generation, and MT. %RNNs are particularly effective in building language models. For example, Mikolov et al. \cite{Mikolov10} developed a statistical language model based on the Elman network \cite{Elman90}.

%Figure \ref{fig: RNN} illustrates the structure of a vanilla RNN. 
% %Therefore, they can captures information from an input sequence, as it reads the sequence, one step at a time. This structure enables RNNs to process sequences of inputs with arbitrary length. 

%Figure
% \begin{figure}[htbp]
% \center
% \includegraphics[bb=0 0 643 207, width=\linewidth]{RNN.png}
% \caption{Network Structure of a Vanilla RNN}
%  \label{fig: RNN}
% \end{figure}

%A RNN consists of a hidden state ${\bf h}$ and an optional output ${\bf y}$. 
Traditional neutral networks are stateless, whereas RNNs have the unique property of being ``stateful''. By reusing the hidden units in the previous layer, RNNs allow cyclically encoding of past information within the networks. 

Let ${\bf x}_i \in \Re^k$  be the $k$-dimensional word vector corresponding to the $i$-th word in a tweet; then, a tweet having n words can be represented as ${\bf X}=({\bf x}_1,...,{\bf x}_T)$. At each time step $t$, the hidden state ${\bf h}_t$ of the RNN is updated as follows:
\begin{equation}
{\bf h}_t=f({\bf h}_{t-1},{\bf x}_t)
\end{equation}
where $f$ is a function that takes a signal ${\bf x}_t$ as input during time step $t$, updates its current state ${\bf h}_t$ based on the influence of ${\bf x}_t$ and the previous state ${\bf h}_{t-1}$. 

\begin{figure}[htbp]
\center
 \includegraphics[bb=0 0 635 484, width=\linewidth]{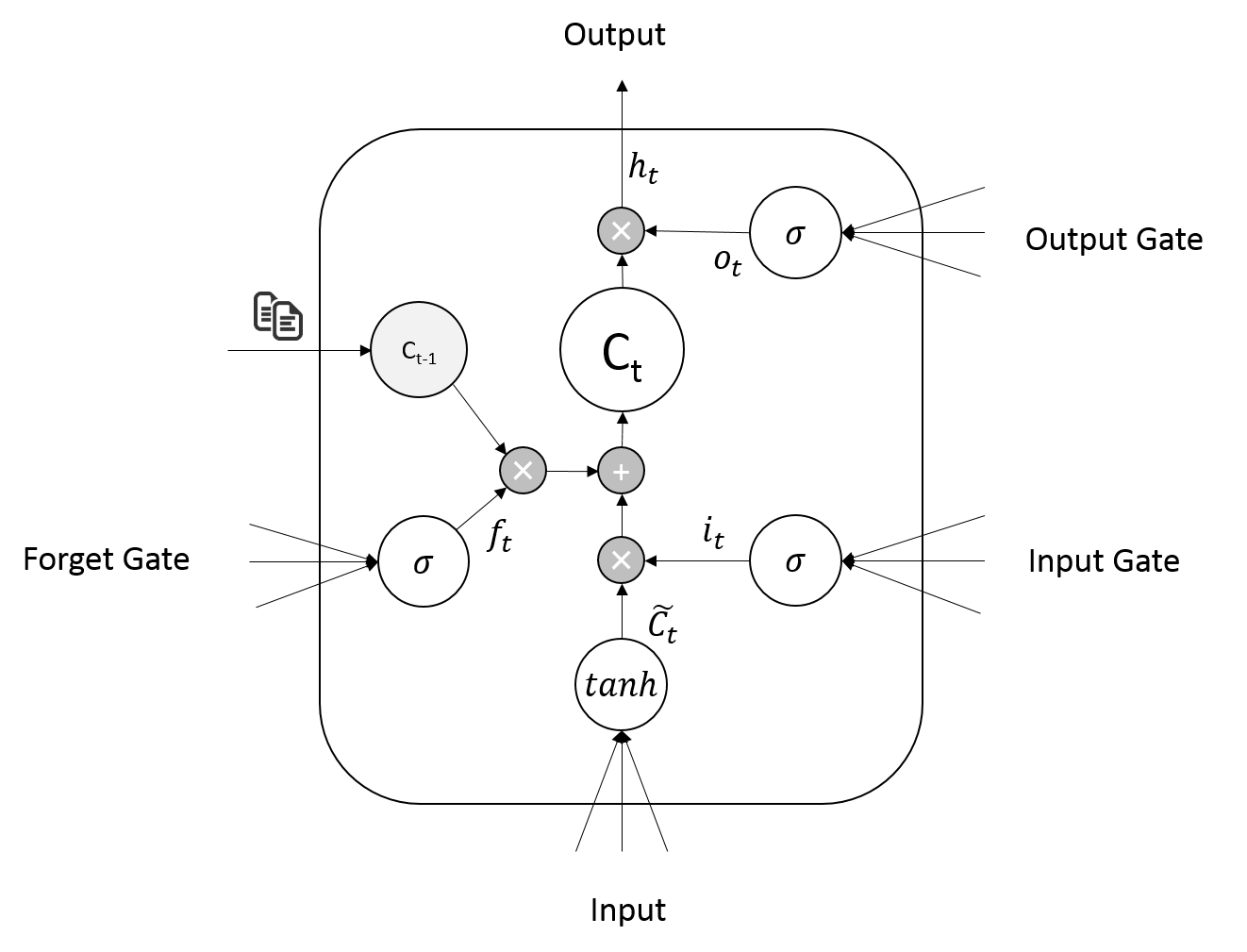}
 \caption{LSTM Memory Block}
 \label{fig: LSTM}
\end{figure}

A vanilla RNN only combines the precious hidden state ${\bf h}_t$ with the current input ${\bf x}_t$, which is not powerful enough to present a complex context. Thus, we used an LSTM  network instead. 
The LSTM model introduces a new structure called a memory block (see Figure \ref{fig: LSTM}). A memory block consists of four main elements: input, output, and forget gates and a self-connected cell. The cell is at the center of the LSTM memory block. Gates can be regarded as water valves, which yield values between 0 and 1, describing how much of each component should be let through. An LSTM memory block has three of these gates, to modulate the cell state. 
 
% Figure

Specifically, the input gate ${\bf i}_t$ controls the candidate state of the cell $\tilde{{\bf C}_t}$; the forget gate ${\bf f}_t$ regulates the previous state of the cell ${\bf C}_{t-1}$; and the output gate ${\bf o}_t$ determined the parts of the cell state ${\bf C}_t$ to output.

Eqs.(\ref{Eq: LSTM1})--(\ref{Eq: LSTM2}) describe how a layer of memory blocks is updated at every time step $t$.
\begin{eqnarray}
\label{Eq: LSTM1}
{\bf i}_t=\sigma({\bf W}_i{\bf x}_t+{\bf U}_i{\bf h}_{t-1}+{\bf b}_i) \\
{\tilde {\bf C}_t} = \sigma({\bf W}_c{\bf x}_t+{\bf U}_c{\bf h}_{t-1}+{\bf b}_c) \\
{\bf f}_t=\sigma({\bf W}_f{\bf x}_t+{\bf U}_f{\bf h}_{t-1}+{\bf b}_f) \\
{\bf C}_t={\bf i}_t*{\tilde {\bf C}_t}+{\bf f}_t*{\bf C}_{t-1} \\
{\bf f}_o=\sigma({\bf W}_o{\bf x}_t+{\bf U}_o{\bf h}_{t-1}+{\bf b}_o) \\
\label{Eq: LSTM2}
{\bf h}_t={\bf o}_t * tanh({\bf C}_t) 
\end{eqnarray}
where ${\bf x}_t$ is the input to the memory block layer at time $t$, ${\bf W}_i$,${\bf W}_c$,${\bf W}_f$,${\bf W}_o$,${\bf U}_i$,${\bf U}_c$,${\bf U}_f$,${\bf U}_o$ are weight matrices, and ${\bf b}_i$,${\bf b}_c$,${\bf b}_f$,${\bf b}_o$ are bias vectors.

Although LSTM memory blocks have a unique (more complicated) way of computing the hidden state, they use the same network structure as the RNN. The lengths of both hidden layer and cell layer for LSTM take the same value as the dimensionality of word embeddings.%In our work, the classification results were decided according to ${\bf y}_T$.
% (compared with Eq.(\ref{Eq: Elman}))

\subsubsection{CNN}
There have been continual debates on which model---the RNN or CNN---is more suited for NLP tasks \cite{Yin17}. Therefore, we use a CNN model for MSA as well.

%Different from RNNs, CNNs have a bionic background. They are known to have been inspired by the human visual cortex\footnote{The visual cortex is a part of the cerebral cortex, which is crucial in processing visual information.}. For example, edge detection, which is a function of the primary visual cortex, can be simulated by applying convolution operation to an image \cite{Wang17}. In addition, although CNNs are designed for image processing, they can be used for NLP tasks. Nevertheless, CNNs for NLP tasks are generally much simpler than those for image processing.

\begin{figure}[htbp]
 \center
 \includegraphics[bb=0 0 459 234, width=\linewidth]{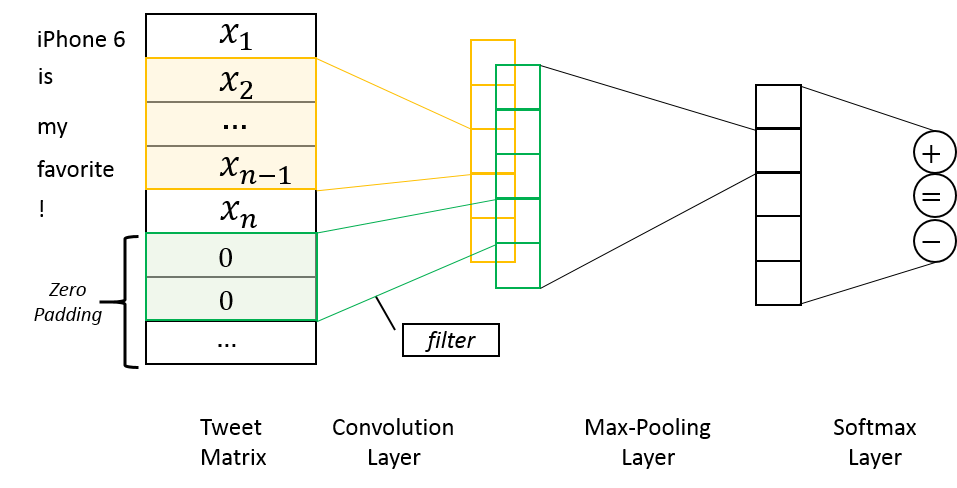}
 \caption{Network Structure of the CNN Model}
 \label{fig: CNN}
\end{figure}

One of the advantages of CNNs is that they have much fewer parameters than fully connected networks with the same number of hidden units, which makes them much easier to be trained. Our CNN is similar to that of Kim \cite{Kim14}. Our CNN model is presented in Figure \ref{fig: CNN}. 

%Figure

As in RNNs, a tweet having n words was represented as follows:
\begin{equation}
{\bf x}_{1:n} = {\bf x}_1 \oplus {\bf x}_2 \oplus  {\bf x}_3 \oplus ... \oplus {\bf x}_n 
\end{equation}
where $\oplus$ is the concatenation operator. Here, the final index of the word vectors in a tweet was $n$ instead of $T$. In general, ${\bf x}_{i:i+j}$ meant the concatenation of words ${\bf x}_i, {\bf x}_{i+1}, ... , {\bf x}_{i+j}$. 

To unify the matrix representation of tweets in different length, the maximum length of all tweets in the dataset was used as the fixed size for tweet matrices. For shorter tweets, zero word vectors were padded at the back of a tweet matrix.

%Typical convolution operations for image processing include identity, edge detection, blur, and sharpening. Different operations can be achieved using different filters, ${\bf w} \in \Re^{hk}$; moreover,  some interesting properties might be discovered by introducing random filters \cite{Wang17}.

The layers of the CNN are formed by a convolution operation followed by a pooling operation. First, we performed a convolution operation to transform a window of $h$ words (i.e., ${\bf x}_{i:i+h-1}$) to generate a feature $c_i$. The procedure was formulated as follows:
\begin{equation}
\label{Eq: CNN1}
c_i = \sigma({\bf w} \cdot {\bf x}_{i:i+h-1} + b) \\
\end{equation}
where ${\bf w}$ denotes a filter map, h is the window size of a filter, $\sigma$ is a non-linear activation function and $b$ is a bias term. 

By applying filter ${\bf w}$ to each possible window of words in a sentence, we obtained a feature map:
\begin{equation}
{\bf c} = [c_1, c_2, ..., c_{n-h+1}]
\end{equation}

Second, we performed a subsampling operation, for which we used the following max-pooling subsampling method based on the idea of capturing the most important feature from each feature map.

\begin{equation}
\label{Eq: CNN2}
c_{max} = max\{{\bf c}\}
\end{equation}

From Eqs.\ (\ref{Eq: CNN1})--(\ref{Eq: CNN2}), a filter generated one $c_{max}$ from a tweet matrix. 

The number of filter maps in our CNN model was 100, and the possible window sizes were $\{3,4, and 5\}$; thus, our model had 300 different filters in total. The corresponding 300 $c_{max}$ formed the penultimate layer, and was then passed to a fully connected softmax layer to predict the global polarity of a tweet.

\section{Experiments} \label{Experiments}
In this section, we compare our deep learning methods with the baseline methods. We first describe our experimental setup, followed by a discussion of the results.

\subsection{Experimental Setup}
\subsubsection{Datasets}
As described in Section \ref{Introduction}, we used the MDSU corpus as our training/test dataset. The MDSU corpus was originally built for deeper sentiment understanding in a multilingual setting; therefore, tweets in it were annotated many fine-grained tags in addition to global (overall) polarity. In this paper, we used global polarities as the classification labels. \cite{Lu17C} filtered out apparent non-emotional tweets and prioritized long tweets with rich language phenomenon during data selection; therefore, the tweets in the MDSU corpus are more complex and longer than those in randomly collected or noisy-labeled tweet datasets.

Table 1 presents the global polarity distribution for each language in the MDSU corpus. The polarity distribution of each language although not perfectly uniform, does not differ largely. Moreover, the polarity distribution of the entire corpus is well-balanced, rendering it a suitable corpus for a three-way sentiment classification. 
The length of a tweet is defined as the number of elements (including words, emoticons, and punctuations) after under-mentioned preprocessing. The maximum length (also the fixed size of the CNN models) of the MDSU corpus is 124: 41 for English, 93 for Japanese, and 124 for Chinese.

%Table
\begin{table}[htbp]
\centering
\small
\setlength{\tabcolsep}{2pt}
%\begin{center}
\caption{Polarity Distribution for Each Language in the MDSU Corpus}
\label{MDSU_distribution}
\begin{tabular}{|c|c|c|c|c|c|c|} \hline
Language & Abbr. &Positive & Neutral & Negative & Total \# & \begin{tabular}[c]{@{}c@{}}Max\\Length\end{tabular} \\ \hline
English & EN  &503 & 526 & 774 & 1803 & 41 \\ \hline
Japanese & JA  &392 & 875 & 534 & 1801 & 93 \\ \hline
Chinese &  ZH &566 & 614 & 638 & 1818 & 124 \\ \hline
Total &  ALL &1461 & 2015 & 1946 & 5422 & 124 \\ \hline
\end{tabular} 
%\end{center}
\end{table}

\subsubsection{Preprocessing}
The language used in social media is more casual than in traditional media. There are many unique ways of expression on Twitter, such as emoticons, Unicode emojis, misspelled words, letter-repeating words, all-caps words, and special tags (e.g., \#, @). These may disturb the learning of word embeddings and classification models; therefore, we preprocessed them to unify the elements in different shapes but with same meanings as much as possible.

For all the three languages, we detected Unicode emojis and replaced them with an ``EMOJI\_CODE'' (e.g., we replaced ``\includegraphics[width=7pt]{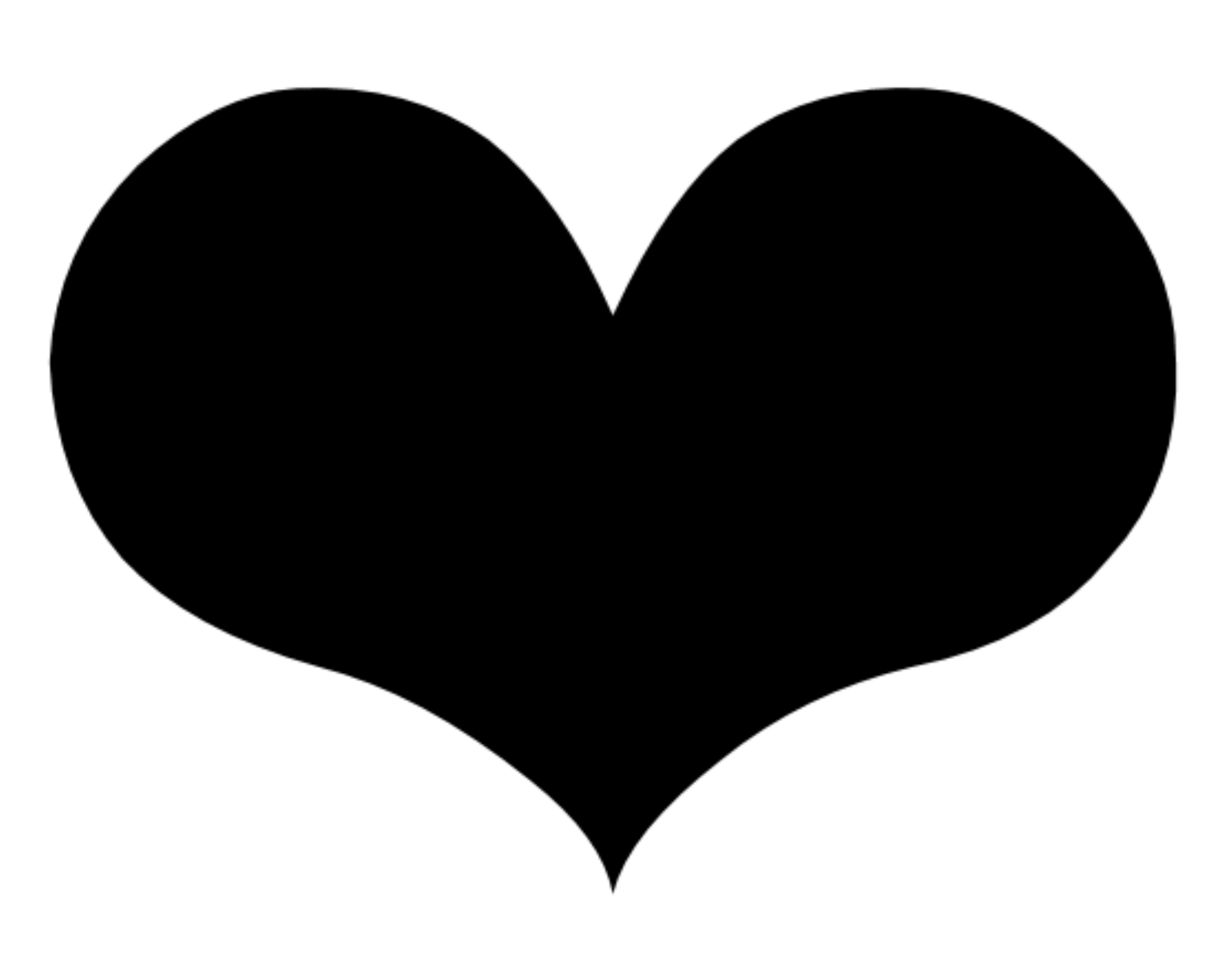}'' with ``EMOJI\_2764''); detected emoticons from easy :-) to complex (((o (*$^{\circ}\bigtriangledown^{\circ}$*) o)))) using regular expressions\footnote{We registered some rare expressions to an ad hoc list.} and replaced them with ``EMOTICON''; and labeled URLs as ``URL''). 

We also performed language-dependent preprocessing. For English, we lowercased English characters and tokenized the tweets with TweetTokenizer\footnote{http://www.nltk.org/api/nltk.tokenize.html}; for Japanese, we normalized Japanese characters and tokenized the tweets with Mecab\footnote{http://taku910.github.io/mecab/}; for Chinese, we transferred traditional Chinese characters to simplified Chinese characters and tokenized the tweets with NLPIR\footnote{http://ictclas.nlpir.org/}.

\subsubsection{Word Embeddings and Translation Matrix}
In addition to the annotated MDSU corpus, we accumulated large collections of raw tweets using Twitter RESTful API by the same query keywords during a one-year period. We first excluded undesirable tweets (e.g., tweets starting with ``RT'') using the same veto patterns as \cite{Lu17C}; then, we checked the preceding 10 tweets to delete the repeating tweets, because similar tweets usually appear in succession. After filtering out the undesirable tweets, the remaining tweets were preprocessed as previously described. The number of remaining tweets was 232,214 (EN), 264,179 (JA), and 148,052 (ZH). The vocabulary size for each collection of tweets was 63,343 (EN), 49,575 (JA), and 52,292 (ZH).

Our vector representation for words was learnt using FastText\footnote{https://github.com/facebookresearch/fastText}. Because the scale of our corpus for word embedding training was relatively small, we set the minimal number of word occurrences as 2. We used the skip-gram model because it generates higher quality representations for infrequent words \cite{Mikolov13X}. The word embeddings for each language were trained separately on its corresponding corpus. Words that were not present in the pre-trained word list were initialized randomly in the deep learning models. 

The dimensionality of our word embeddings was 100, and the Japanese/Chinese spaces were transformed by their respective translation matrices. For the translation matrix, we set k as 3500, which implied that the top 3500 English words and their translations were the pivot word pairs. We split the 3500 pivot word pairs into two sets---training set (3000 words) and test set (500 words). The translation matrices were obtained based on the training sets. As a validation, we calculated the change of Euclidean/cosine distances for each word pair in the test set before and after the mapping; Table \ref{Distance Sum} depicts the decrease in the sum of the two distances.

%Table
\begin{table}[htbp]
\centering
\small
\setlength{\tabcolsep}{2pt}
%\begin{center}
\caption{Sum of Embedding Distances of Word Pairs in the Test Set}
\label{Distance Sum}
\begin{tabular}{|c|c|c|c|} \hline
\multicolumn{2}{|c|}{Language} & Before Mapping & After Mapping \\ \hline
\multirow{2}{*}{Japanese} 
& {\scriptsize Euclidean Distance} & 2,455.94  & 2,135.51 \\ \cline{2-4}
& {\scriptsize Cosine Distance} & 458.37 & 440.82 \\ \hline
\multirow{2}{*}{Chinese} 
& {\scriptsize Euclidean Distance} & 2,457.05 & 2,098.76 \\ \cline{2-4}
& {\scriptsize Cosine Distance} & 496.01 & 490.88 \\ \hline
\end{tabular} 
%\end{center}
\end{table}

\subsubsection{\bf Model Hyper-parameters}
All the methods were tested using 10-fold cross validation. For the deep learning models, we randomly selected 10\% of the training splits of cross-validation as the developed datasets to tune parameters for an early stopping. 

For fair comparison, we empirically set the hyper-parameters for deep learning models as consistent as possible. Both trainings were completed using a stochastic gradient descent (SGD) algorithm for shuffled mini-batches with the Adadelta update rule, with a mini-batch size of 50. The dropout technique is effective in preventing co-adaptation of hidden units by randomly setting a portion of the hidden units to zeroes during feedforward/backpropagation. Therefore, to prevent overfitting, we employed the dropout technique for both deep learning models on their penultimate softmax layers, with a dropout rate of 0.5. We did the same for the dimensionality of word embeddings; the lengths of both the hidden and cell layer for LSTM were 100. 

\subsection{Result and Discussion}
\subsection{Baselines}
Table \ref{baseline results} presents the classification accuracies of baselines. 

According to Table \ref{baseline results}, the average accuracy of separate SVM classifiers over original datasets was the same as it over translated datasets. This showed that the same method did not necessarily perform worse after being translated by MT for monolingual datasets. In addition, the performance of MT+SVM model (use all translated tweets) was worse than the average accuracy of separate SVM classifiers over original datasets (53.0\% vs. 54.5\%), showing the limitation of traditional paradigm(i.e., ``MT + machine learning''). 

For classifiers directly used the cumulation of unigram and bigram, both SVM and Banea (2010)* performed better than MT+SVM by 0.8\% and 3.4\%, respectively. The increases indicate that the use of cumulation of n-gram is effective; although this may result in the problem of data sparseness \cite{Banea10}, it could be mitigated by feature selection.

%Table
\begin{table}[htbp]
\centering
\small
\setlength{\tabcolsep}{5pt}
%\begin{center}
\caption{Results of Baselines}
\label{baseline results}
\begin{tabular}{|c|c|c|c|} \hline
\multicolumn{1}{|c|}{Model} & \multicolumn{1}{c|}{Dataset} & \multicolumn{1}{c|}{Feature} &\multicolumn{1}{c|}{Accuracy} \\ \hline \hline
Average & -- & -- & 0.545 \\ \hline
{\scriptsize \quad SVM} & {\scriptsize EN} &  {\scriptsize unigram+bigram} & {\scriptsize 0.529} \\
{\scriptsize \quad SVM} & {\scriptsize JA} &  {\scriptsize unigram+bigram} & {\scriptsize 0.596} \\
{\scriptsize \quad SVM} & {\scriptsize ZH} &  {\scriptsize unigram+bigram} & {\scriptsize 0.509} \\ \hline
Average & -- & -- & 0.545 \\ \hline
{\scriptsize \quad SVM} & {\scriptsize EN} & {\scriptsize unigram+bigram} & {\scriptsize  0.529} \\
{\scriptsize \quad SVM} & {\scriptsize Translated JA} & {\scriptsize unigram+bigram} & {\scriptsize 0.591} \\
{\scriptsize \quad SVM} & {\scriptsize Translated ZH} & {\scriptsize unigram+bigram} & {\scriptsize 0.515} \\ \hline
MT+SVM & {\scriptsize Translated ALL} & {\scriptsize unigram+bigram} & {\bf 0.530} \\ \hline  \hline
SVM & {\scriptsize ALL} & \begin{tabular}[c]{@{}c@{}} {\scriptsize cumulation of} \\ {\scriptsize unigram+bigram} \end{tabular}& 0.538 \\ \hline
Banea (2010)* & {\scriptsize ALL} & \begin{tabular}[c]{@{}c@{}} {\scriptsize cumulation of} \\ {\scriptsize unigram+bigram} \end{tabular} & {\bf 0.564} \\ \hline
\end{tabular} 
%\end{center}
\end{table}

\subsection{Deep Learning Methods}
Table \ref{DL results} presents the classification accuracies of the deep learning models; the input of word embeddings for the models in this Table involved no transformation.

First, our deep learning paradigm performs better than the MT+SVM method (traditional paradigm). Specifically, parameter-sharing LSTM and CNN models outperformed MT+SVM model by 1.2\% and 4.3\%, respectively. Thus, the deep learning paradigm is more efficient than the traditional paradigm. In addition, the LSTM performed worse than the Banea (2010)* baseline, whereas the CNN excelled. Thus, CNN is more suitable for MSA than LSTM.

Besides, we also conducted the learning separately on each language split. The results revealed that the average accuracies of separate LSTM/CNN classifiers were a little higher than the accuracy of the mixed case (54.4\% vs. 54.2\%, and 58.1\% vs. 57.3\%), implying that the deep learning methods did not improve after using the entire dataset. This was a result of the heterogeneity of vector spaces of word embeddings, because the raw word embeddings were learned separately.

Furthermore, we observed that both MT + LSTM and MT + CNN models (trained on the translated datasets and using only English word embeddings) performed worse than the LSTM and CNN models (trained on the original datasets and using multilingual word embeddings). Ideally, if JA/ZH were perfectly translated, the performance should have increased. This suggests that the noises that MT brings in are greater than the heterogeneity of multilingual word embeddings does.

%Table
\begin{table}[htbp]
\centering
\small
\setlength{\tabcolsep}{5pt}
%\begin{center}
\caption{Results of Deep Learning Models}
\label{DL results}
\begin{tabular}{|c|c|c|} \hline
\multicolumn{1}{|c|}{Model} & \multicolumn{1}{c|}{Dataset} & \multicolumn{1}{c|}{Accuracy} \\ \hline \hline
Average & -- & 0.544 \\ \hline 
{\small \quad LSTM} & {\small EN} &  {\small 0.531} \\ 
{\small \quad LSTM} & {\small JA} &  {\small 0.569} \\ 
{\small \quad LSTM} & {\small ZH} &  {\small 0.532} \\ \hline 
MT+LSTM & Translated ALL & 0.541 \\ \hline
Parameter-sharing LSTM & ALL & {\bf 0.542} \\ \hline \hline
Average & -- & 0.581 \\ \hline
{\small \quad CNN} & {\small EN} & {\small 0.578} \\ 
{\small \quad CNN} & {\small JA} & {\small 0.610} \\ 
{\small \quad CNN} & {\small ZH} & {\small 0.553} \\ \hline
MT+CNN & Translated ALL & 0.564 \\ \hline
Parameter-sharing CNN & ALL & {\bf 0.573} \\ \hline
\end{tabular} 
%\end{center}
\end{table}

\subsection{Deep Learning Methods using Transformed Word Embeddings}
The unification of different vector spaces was expected to further improve the deep learning paradigm. Table \ref{TM results} presents the classification accuracies of the deep learning models before and after space coordination. According to Table \ref{TM results}, the effectiveness of LSTM and CNN models were divided. We observed that after space transformation, the accuracy of LSTM decreased by 0.6\%, whereas the accuracy of CNN increased by 1.4\%. This suggests that the same vector space transformation does not necessarily suitable for different kinds of network structures.

Overall, the performance of the CNN model fed with transformed word embeddings was most effective.

%Table
\begin{table}[htbp]
\centering
\small
\setlength{\tabcolsep}{3.6pt}
%\begin{center}
\caption{Results of Deep Learning Models Before and After Space Transformation}
\label{TM results}
\begin{tabular}{|l|c|c|c|} \hline
\multicolumn{1}{|c|}{Model} & \multicolumn{1}{c|}{Dataset} & \multicolumn{1}{c|}{Word Embedding} & \multicolumn{1}{c|}{Accuracy} \\ \hline \hline
\multirow{2}{*}{\begin{tabular}[c]{@{}c@{}}Parameter-sharing\\LSTM\end{tabular}} 
& ALL & Raw (Table \ref{DL results}) & {\bf 0.542} \\ \cline{2-4}
% & ALL & D300-D100 & 0.538 \\ \cline{2-4}
& ALL & Transformed & 0.536 \\ \hline \hline
\multirow{2}{*}{\begin{tabular}[c]{@{}c@{}}Parameter-sharing\\CNN\end{tabular}}
& ALL & Raw (Table \ref{DL results}) & 0.573 \\ \cline{2-4}
% & ALL & D300-D100 & 0.580 \\ \cline{2-4}
& ALL & Transformed & {\bf 0.587} \\ \hline
\end{tabular} 
%\end{center}
\end{table}

\section{Conclusion and Future Work} \label{Conclusion}
In this paper, we proposed a novel deep learning paradigm for MSA. We map monolingual word embeddings into a shared embedding space, and used parameter-sharing deep learning models to unify the processing of multiple languages. The tests on a well-balanced tweet sentiment corpus---the MDSU corpus---revealed the effectiveness of our deep learning paradigm. Especially, our CNN model fed with translation matrix-transformed word embeddings achieves a rise of 2.3\%, comparing with the strong Banea (2010)* baseline. 

Our paradigm provides a great cross-lingual adaptability.  Training tweets in any other language can be transferred into vector representation using transformed word embeddings, and then combined with the learning process of the deep learning models. 

The novelty of our study is not in the complexity of the network itsel, but more in the unification of heterogeneous monolingual word embeddings and the parameter-sharing model for multilingual datasets. In the future, we plan to attempt more complex transformation methods and network structures. 

%Moreover, pre-trained monolingual word embeddings can be further tuned using word-level polarities of words in the context that provided in the MDSU corpus. Finally, unsupervised text tokenizers, such as SentencePiece\footnote{https://github.com/google/sentencepiece}, may liberate us from using any language-specific tokenizers, which makes the proposed paradigm for MSA completely language-independent.

\bibliography{refs}
\bibliographystyle{ijcnlp2017}

\end{document}